\documentclass[letterpaper, 10 pt, conference]{ieeeconf}  %

\IEEEoverridecommandlockouts                              %

\usepackage{graphicx} 
\usepackage{amsmath} %

\usepackage{float}
\usepackage{algorithm2e}[ruled]
\usepackage{caption}
\usepackage{subcaption}
\usepackage{fixltx2e}
\usepackage{xcolor}
\usepackage{soul}

\title{\LARGE \bf
Leveraging Foundation Models To learn the shape of semi-fluid deformable objects.%
}

\author{Omar El Assal$^{1,2}$, Carlos M. Mateo$^{3}$, Sebastien Ciron$^{1}$ , David Fofi$^{2}$%
\thanks{$^{1}$Omar El Assal And Sebastien Ciron are with Alstom transports, 105 allée Albert Einstein – 71200 LE CREUSOT – France,
        {\tt\small omar.el-assal@alstomgroup.com}, and  {\tt\small sebastien.ciron@alstomgroup.com} }%
\thanks{$^{2}$David Fofi and Omar El Assal with the IMVIA EA 7535 laboratory, university of Burgundy,
         720 Av. de l'Europe, 71200 Le Creusot, France
        {\tt\small david.fofi@u-bourgogne.fr}}%
\thanks{$^{3}$Carlos M. Mateo with the ICB UMR CNRS 6303, Universite de Bourgogne,
        9 Av. Alain Savary, 21000 Dijon, 
        {\tt\small carlos-manuel.mateo-agullo@u-bourgogne.fr}}%
}

\begin{document}

\maketitle
\thispagestyle{empty}
\pagestyle{empty}

\begin{abstract}
    One of the difficulties imposed on the manipulation of deformable objects is their characterization and the detection of representative keypoints for the purpose of manipulation.
    A keen interest was manifested by researchers in the last decade to characterize and manipulate deformable objects of non-fluid nature, such as clothes and ropes.
    Even though several propositions were made in the regard of object characterization, however researchers were always confronted with the need of pixel-level information of the object through images to extract relevant information.
    This usually is accomplished by means of segmentation networks trained on manually labeled data for this purpose.
    In this paper, we address the subject of characterizing weld pool to define stable features that serve as information for further motion control objectives.
    We achieve this by employing different pipelines. The first one consists of characterizing fluid deformable objects through the use of a generative model that is trained using a teacher-student framework. 
    And in the second one we leverage foundation models by using them as teachers to characterize the object in the image, without the need of any pre-training and any dataset.
    The performance of knowledge distillation from foundation models into a smaller generative model shows prominent results in the characterization of deformable objects. The student network was capable of learning to retrieve the keypoitns of the object with an error of 13.4 pixels. And the teacher was evaluated based on its capacities to retrieve pixel level information represented by the object mask, with a mean Intersection Over Union (mIoU) of 75.26\%.
\end{abstract}

\section{INTRODUCTION}

In order to accomplish manipulation tasks successfully, a robot has to perceive and comprehend the manipulated object. %
This consists of analysing information from several sources like images, points clouds or the model of the object~\cite{c27,c28,c29}.
Most of the research found in the literature focused on modeling and characterising rigid bodies~\cite{c26}. 
However, the ever-increasing need for robots to interact with day-day objects, alongside the recent advances in machine learning and computer vision have raised an interest in the subject of deformable objects.  
For instance, the topic of modeling and estimating the state of these objects have attracted the interest of several researchers~\cite{c1,c19,c24,c25,c37}.
Which indicates that the manipulation of deformable objects presents several challenges imposed by the deformations of the manipulated object~\cite{c10}.
This paper targets fluid-like objects of deformable nature, and that presents high-dynamical topological surface changes. A texture rarely discussed in state of the art deformable objects manipulation of fabric and ropes ~\cite{c10,c8,c9,c12}.
The shape of these fluid bodies is often governed by complex physical modeling, and their shape is difficult to predict. For instance, weld pools are fluid deformable objects, often described as visco-elastic, and their shape is influenced by different process parameters: robot speed, welding position, and the type of welded geometry~\cite{c121, c122}.

\begin{figure}[t]
   \centering
   \includegraphics[width=\linewidth]{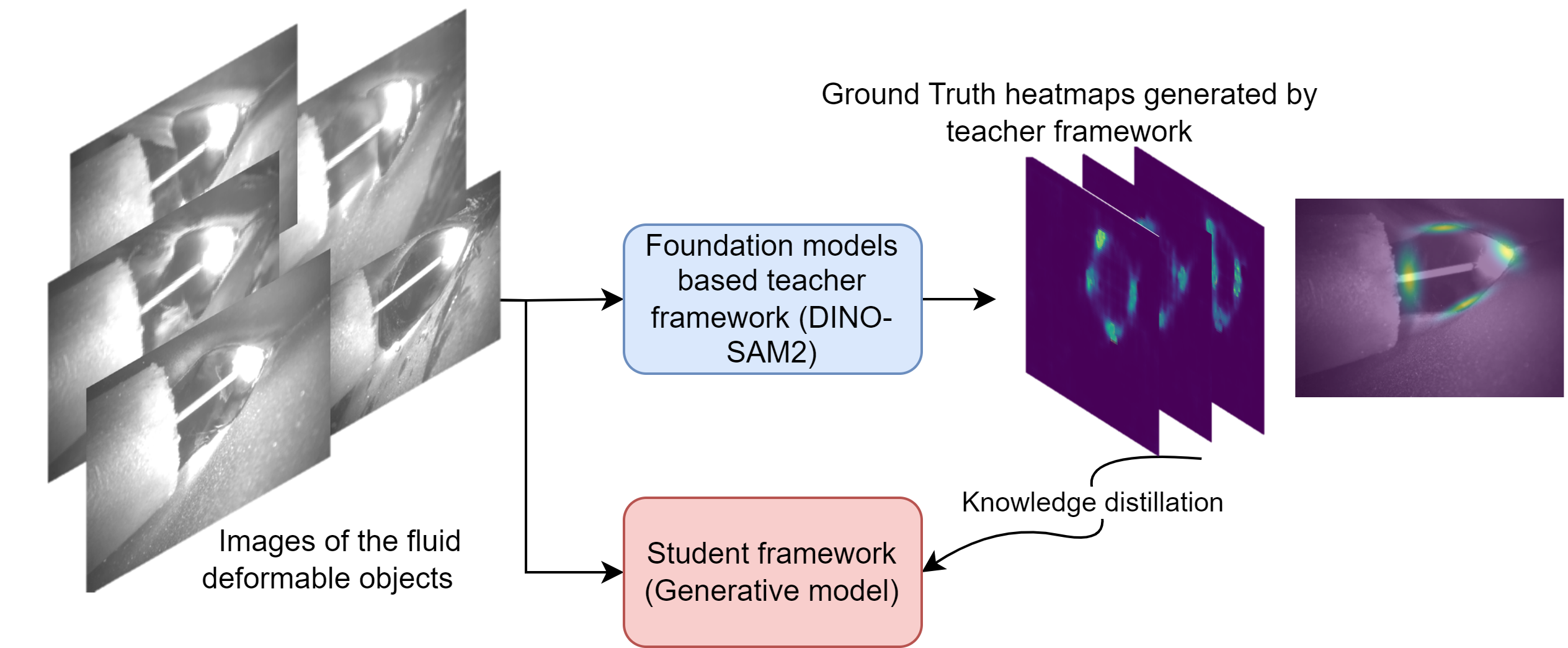}
   \caption{Teacher-Student framework adopted for training a student network to characterize fluid deformable objects.}
   \label{fig:teacher-student}
\end{figure}

   \begin{figure*}[t]
      \centering
      \includegraphics[width=0.8\textwidth]{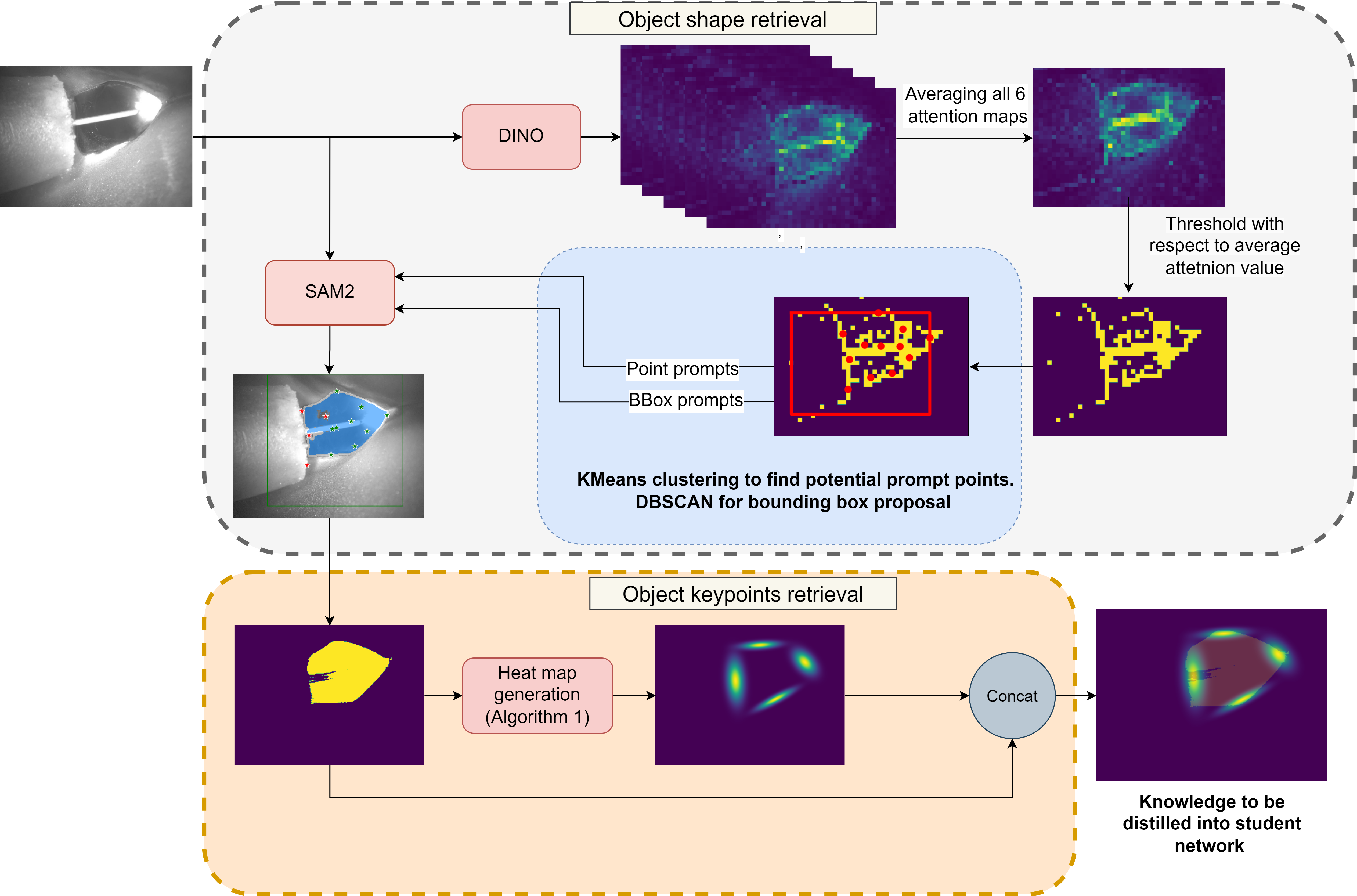}
      \caption{Framework of the proposed DINO-SAM method}
      \label{framework}
   \end{figure*}

We tackle the issue of characterizing fluid deformable objects by describing them according to the shape of their contours.
Our approach is based on a Teacher-student framework in which the knowledge of a teacher that consists of foundation models is distilled into a student lighter generative network as shown in Fig.~\ref{fig:teacher-student}.
The teacher network consists of  two foundation models, DINO~\cite{1223} and SAM2~\cite{1224} to retrieve the shape of the object.
We describe the object through heatmaps representing the keypoints of the object.
These heatmaps are first generated by the teacher, and the knowledge is later transferred to the student network.
The proposed approach targets an industrial application which is the state and shape estimation of weld pool in robotic welding task.
Because modelling the deformations of molten metal pool is a key element to perform advanced control of the process such as feature based visual servoing.
We also test and validate our proposal on this task and on a real-world dataset, generated in harsh industrial conditions.

\section{Related work and motivation}

Describing deformable objects have always been a research question that accompanied the manipulation of these objects~\cite{c34}; In many cases they were represented as objects as continuous geometrical entities like surfaces and curves~\cite{c13,c14}. 
Some researchers however described them in a discrete representation through meshes, skeletons or landmark points~\cite{c2}.
And these objects were of different shapes and natures. In some cases they were of a linear deformable objects nature, and their characterization serves as a feedback for for later use in a geometric optimal control problem~\cite{c23}.
In other cases they were of more complex shapes to represent such as clothes ~\cite{c6}.
The problematic of deformable clothing grasping in this case was resolved by assimalting clothes to polygonal shapes described by their corners.
From a different perspective,~\cite{c5} relied on the detection of extremed edges and points to represent objects such as towels and ropes. They presented a method that is based on the use of Convolutional Neural Networks. While this approach worked well for fabric, it still does not take into account the deformable nature of fluid objects.
In ~\cite{c7}, Coherent Point Drift (CPD) is applied after the points are registered with a Gaussiant Mixture model (GMM), to maintain feature from one frame to another.
Softer objects such as dough characterization and manipulation have also been studied by researchers, in~\cite{c35} visual images with depth maps are analyzed by Graph Neural Netwroks to learn a particle-based model of the object.
Besides that,  tissues are another type of soft deformable objecs that were explored by researchers. Feature extraction and state tracking of deformable of soft tissues by deep neural networks approach exist in the literature ~\cite{c36}. 
Nevertheless, manipulation and characterization of softer deformable objects, such as fluid and fluid like objects have been introduced~\cite{c3,c1}.
Where in~\cite{c3} the manipulation of semi-fluid objects in a wok is addressed. These objects are considered as a single particle represented by the center of the object. 
In~\cite{c1} a model-based method was used to manipulate water.

As discussed in earlier sections, the methods present in the state of the art do not apply to fluid and semi-fluid deformable objects. Although these methods are efficient for their specific task, they fail to be generalized for other types of applications and objects, since many of the proposed solutions are application oriented ~\cite{c11}.
Providing model-free methods to estimate the shape and to characterize deformable objects is the main motivation of this work. This motivation is backed by the unpredictable nature of semi-fluid objects, such as weld pool and glue, and the difficulties of predicting their deformations.
To address the problem of generalization, we propose the use of foundation models. Foundation models are large-scale models trained on a large scale of data and tasks, which allows for better analysis and understanding of this data \cite{foundation}.
We use these models as teachers for a smaller student network as depicted in Fig.~\ref{fig:teacher-student}.

As for the student network, we use a generative model. Generative models are mathematical models that analyze the underlying patterns in the streams of data.
They have the capacity of generating new data with similar characteristics by learning underlying representations~\cite{c15}.
Several architectures are present in the literature such as Generative Adversarial Networks~\cite{c30}, Auto-encoders architectures~\cite{c31}, Gaussian Mixture Models (GMM), and Hidden Markov Models (HMM))~\cite{c15}.
As for visual image generation the most commonly used are diffusion models~\cite{c32}, GANs~\cite{c30}, auto-encoders~\cite{c31} and Variational Auto-encoders~\cite{c4}. These models often need heavy computing powers and have complex convergence criteria~\cite{c16}.
Amongst these models, the lighter, the fastet to covnerge and the most convenient at learning latent representation are Variational Autoencoders. The only cost is the clarity of the output data~\cite{c17}.
Thus, 
we define the architecture of the student network to be Variational Auto-Encoder (VAE). 
    
\section{Feature points extraction using Generative Models}
    
We employ the teacher-student architecture of Fig.~\ref{fig:teacher-student} in our framework. We consider the output of the teacher to be the corresponding ground truth.
In this framework, the teacher takes as input the corresponding image of the deformable object and outputs a heatmap describing its shape. This heatmap is stacked with the mask of the object in the image to represent the output of the teacher. On the other hand, the student network has a variational auton-encoder architecture with a ResNet backbone.
As illustrated in Fig.~\ref{framework}, the teacher consists of two different pipelines. The first one consists of finding the mask of the object by combining the two foundation models discussed earlier, DINO and SAM2, and the second one is dedicated to generate relevance heatmaps describing the keypoints of the object. 
In first place, the image is passed through DINO model. DINO is a self distillation network that is capable of generating attention maps describing the observed scene. After retrieving the six attention maps, their average is thresholded with respect to the mean value of attention to propose a 2D array with highest attention values.
After that, we employ a pipeline to extract SAM2 prompts. SAM2 is another prompt based foundation model, that takes as input an image and possible prompts, and proposes a possible mask for these prompts on the output.
We pass the original image through the SAM2 model, alongside with the extracted prompts from the DINO attention map in order to retrieve the pixel-level information of the object represented by the mask.
After that, a heatmap that describes the keypoints of the object is proposed by Algorithm ~\ref{alg:normal_lines}, this heatmap represented the most probable location of the keypoints, according to the normal lines on the contour of the object.
Subsequently, the extracted heatmap alongside the previously defined mask are concatenated to represent the ground truth label which is distilled into the student network.

\begin{figure}[t]
   \centering
   \includegraphics[width=0.95\columnwidth]{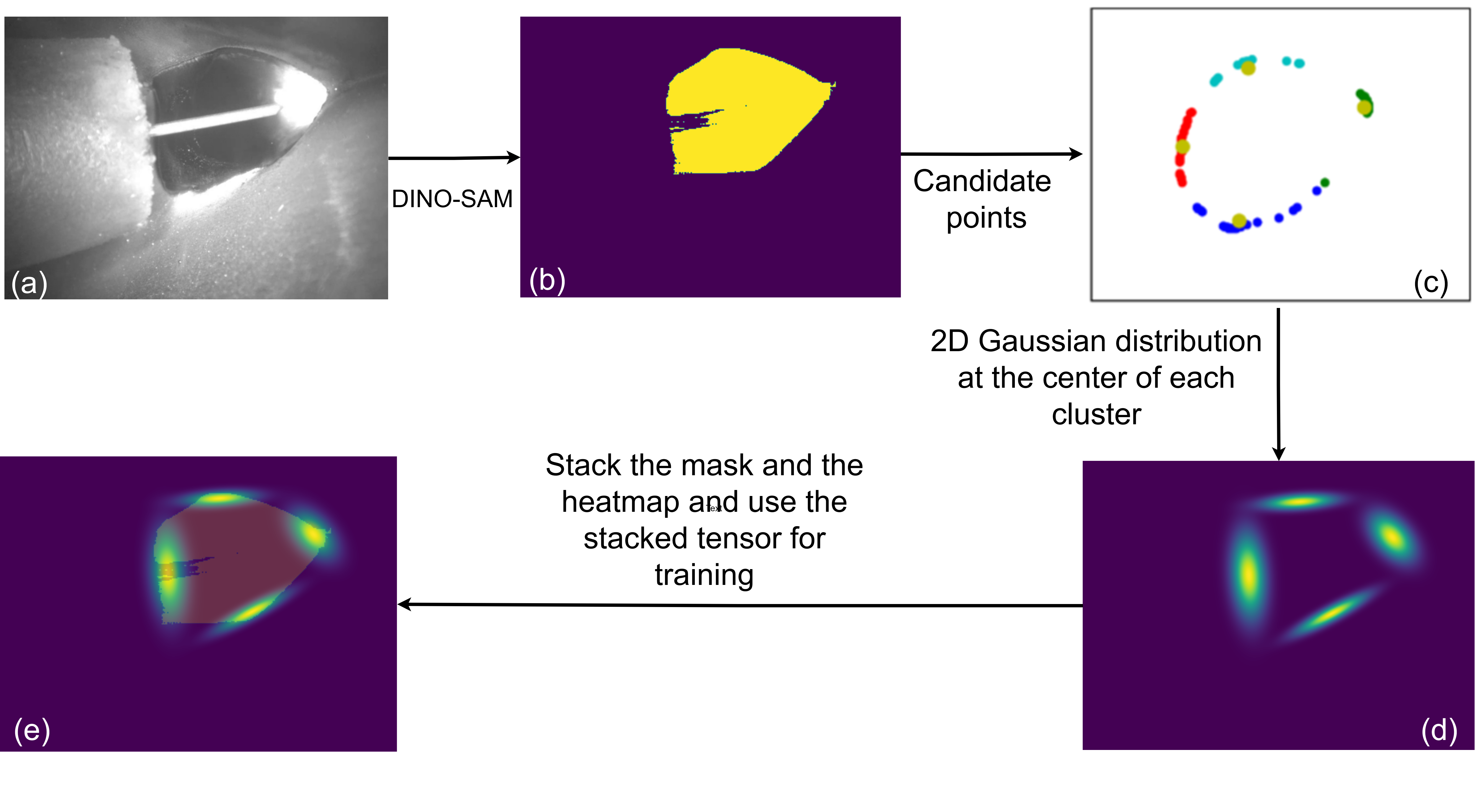}
   \caption{The phases of finding the Gaussian heatmap for an image. \textbf{\emph{(a)}} Initial image. \textbf{\emph{(b)}} Output of DINO-SAM. \textbf{\emph{(c)}} potential candidates clustered into 4 clusters. \textbf{\emph{(d)}} The Gaussian heatmap at the center of each cluster. \textit{\textbf{(e)}} Output of teacher : mask and heatmap.}
   \label{heatmap_extraction}
\end{figure}

The generation of the heatmap goes through the different steps that are highlighted in Algoirhtm ~\ref{alg:normal_lines} and depcited in Fig~\ref{heatmap_extraction}.
The contour of the object is extracted from the output mask of DINO-SAM. 
And Sobel operator is applied to find the values of normal lines. 
We refer to these lines according to their angle $\theta$.
Let $C$ be the set of these points $p_i$,
\begin{equation}
    \left[\left(x_i, y_i\right), \theta_i\right] = p_i
\end{equation}

\noindent the pair $(x_i, y_i)$ represents the coordinates of each point $p_i$ in the contour $C$ of the object, and $\theta_i$ the corresponding angular coordinate, codifying its normal angle.

The difference between $\theta_i$ and $\theta_{i-1}$ is calculated and filtered to find the points at which the variation is the highest.
The set of found points $P$ is denoted as,
\begin{equation}
    \begin{split}
        P = \{(x_i,y_i) \quad | \quad |\theta_i - \theta_{i-1}| > \lambda\}    
    \end{split}
\end{equation}

\noindent $\lambda$ is the threshold value for filtering out low frequency variations.
$K$-means is later applied to cluster the points of $P$ into $k$ groups.  And each of these groups $P_j$ is a region of interest described by a 2D Gaussian distribution, of center $\mu$ and covariance matrix $\Sigma$

\noindent %
Following the clustering of the points and the calculation of the covariance matrices, the heatmap $F_j$ of cluster $P_j$ is defined at the center of this cluster, %
Thus the ground truth heatmap is finally computed as the mixture of all $F_j$,

\begin{equation}
    F = \sum^{k}_{j=1}{F_j}
\end{equation}

\noindent Algorithm~\ref{alg:normal_lines} highlights the steps of the heatmap calculation discussed before.
\RestyleAlgo{ruled}
\begin{algorithm}
   \caption{Heatmap calculation}\label{alg:normal_lines}
   \KwData{$I$ and $\lambda$. Respectively, the input image and a threshold parameter
   }
   \KwResult{$F$ a matrix of the shape (W,H,2); first channel is the heatmap, second channel is the mask}
    $S = Block\_B(I)$\;
    $C = findContour(S)$\;
    $\nabla S_x = sobel_x(S)$\;
    $\nabla S_y = sobel_y(S)$\;
    $\Theta=\{\}$\;
    \ForEach { $p_i \in C$}{
    $\theta_i = \nabla S_x \otimes \nabla S_y$\;
    $\Theta = \Theta \cup \{\theta_i\}$\;
    }
    $P=\{\}$\;
    \ForEach{$\theta_i \in \Theta$}{
        \If{$|\theta_i - \theta_{i-1}| > \lambda$}{
            $P = P \cup \{ (x_i, y_i) \}$
        }
    }
    $P_j = Kmeans(P)$\;
    $F=\{\}$\;
    \ForEach{$P_j$}{
     $\Sigma_j = covMatrix(P_j)$\;
    $F += f_j(P_j; \mu_j, \Sigma_j)$
    } \
    \Return{$F\otimes S$}
\end{algorithm}

The output of the teacher framework, Fig.~\ref{heatmap_extraction}e, represents the ground truth knowledge that is distilled into the ResNet-VAE stduent~\cite{c22,c4,c21}. This type of networks encodes the input into a latent probabilistic space of continuous nature.

In our application, we encode the image of the weld pool using resnet~\cite{c21}, and we decode the probabilistic latent vector to reconstruct the heatmap and the mask using a fully connected convolutional layer, outputing a two channels matrix of shape (W,H,2). Here, W and H are the dimension of the input image, and each of the channels represent respectively the DINO-SAM mask and the calculated heatmap of Fig.\ref{heatmap_extraction}

The purpose of reconstructing the mask of the image instead of the image itself using the student network (the VAE), is that reconstructing a single element of the image instead of all three channels helps the network focus on a single zone and a single shape, which helps converging the network and accelerates it.
t

\section{Data acquisition and dataset}
\begin{figure}[h]
    \centering
    \includegraphics[width=0.9\linewidth]{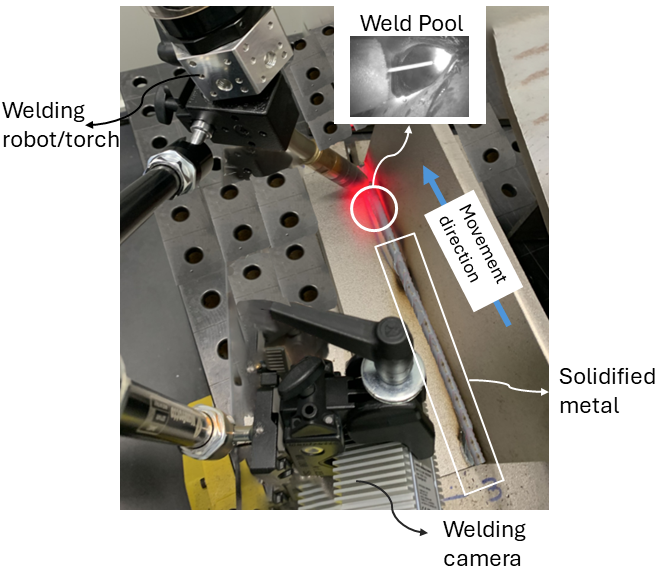}
    \caption{The setup used for dataset generation (arc-off)}
    \label{fig:data_generation_config}
\end{figure}

An industrial application that consists of characterizing the shape of weld pool as taken as case in point.
Characterizing weld pool is important to control the robotized welding process and to analyze the quality of the weld.
Since no existing benchmark or dataset exist for that purpose we create our own dataset by equipping a camera to an industrial robot, equipped with a welding source.
The changes in geometry, robot speed, and process parameters accompanied with external perturbations impose deformations on the molten pool manipulated by the robot.
Nine video sequences are taken of the weld pool in these conditions and by welding actual samples in a fillet joint configuration (Tee joint).
Fig.~\ref{fig:data_generation_config} shows the different elements of the welding setup.
Table~\ref{table:dataset} shows the different parameters used to acquire the dataset.
\begin{table}[h]
    \caption{The conditions of dataset acquisitions. In addition to these conditions, positional offset is applied.}
    \label{table:dataset}
    \scalebox{0.9}{\begin{tabular}{|c|c|c|c|c|}
        \hline
        \multicolumn{1}{|l|}{\begin{tabular}[c]{@{}l@{}}Wire feed rate \\ (10 m/min)\end{tabular}} & \begin{tabular}[c]{@{}c@{}}Voltage\\ (V)\end{tabular}                  & \begin{tabular}[c]{@{}c@{}}Current\\ (A)\end{tabular}                  & \begin{tabular}[c]{@{}c@{}}Arc Length \\ Correction(\%)\end{tabular} & \begin{tabular}[c]{@{}c@{}}Robot Velocity\\ (cm/min)\end{tabular} \\ \hline
        10                                                                                         & 31.9                                                                   & 328                                                                    & \textbf{1.5/3.5/4.2}                                                      & 30                                                                \\
        \hline
        10                                                                                         & 31.9                                                                   & 309                                                                    & 4                                                                    & \textbf{30/35/40 }                                                         \\
       \hline
       \textbf{ 9.5/9.8/10 }                                                                               & \begin{tabular}[c]{@{}c@{}}Synergic\\ laws\end{tabular} & 
       \begin{tabular}[c]{@{}c@{}}Synergy lines\\ laws\end{tabular} & 4                                                                    & 30                                                                \\
        \hline
    \end{tabular}}
\end{table}

\begin{figure}[h]
    \centering
    \includegraphics[width=\linewidth]{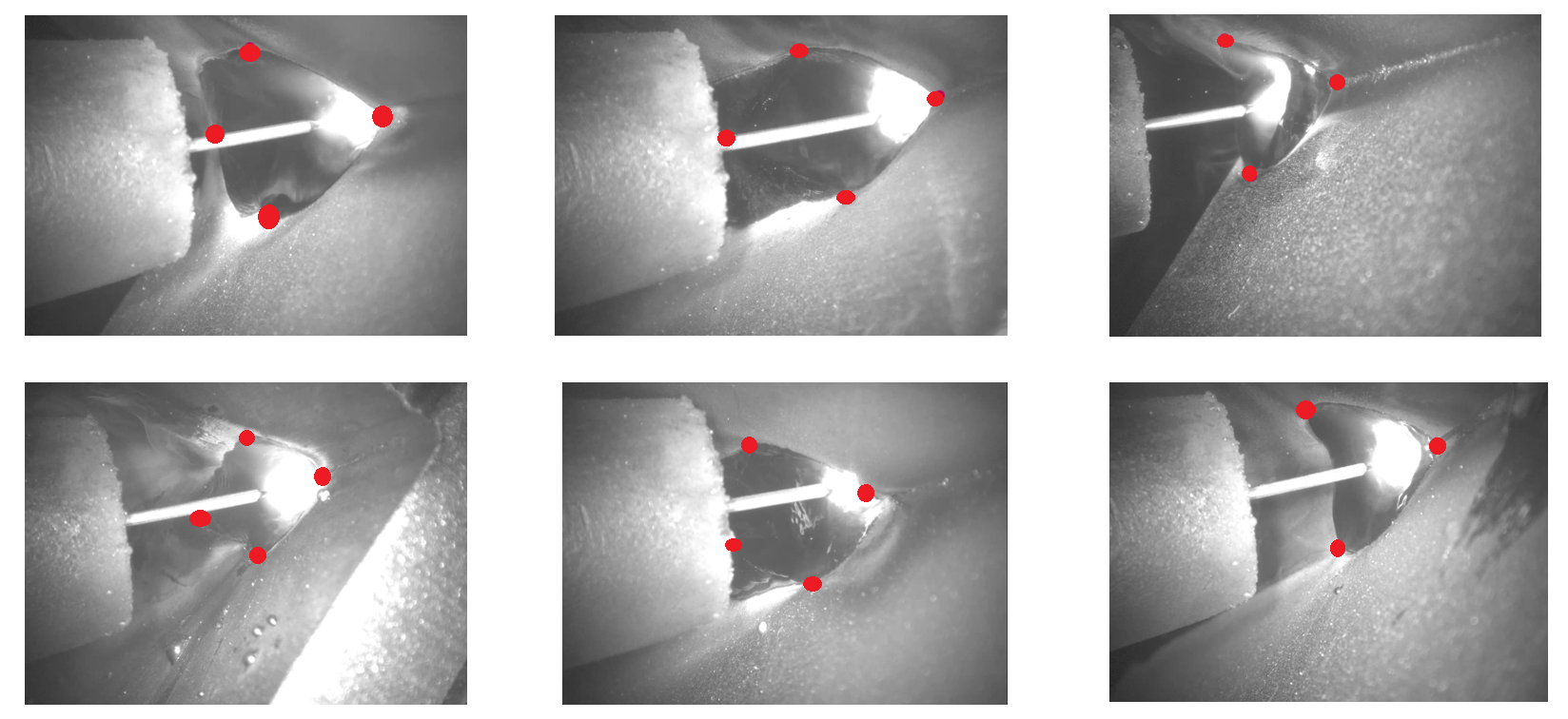}
    \caption{Examples of the different shapes encountered in the dataset. In red are the features that should be used to control the robot}
    \label{fig:different_shapes}
\end{figure}

This dataset is manually labeled with the possible keypoints that represent the shape of the weld pool, to validate the performance of the student network.
A subset of the acquired sequences are labele for segmentation task, for the purpose of validating the performance of the DINO-SAM2 framework for object shape retrieval. Some examples of the shapes encountered in the dataset are shown in Fig.~\ref{fig:different_shapes}
\section{Results}
The performance of the teacher network is validated according to its capacity of retrieving accurate masks of the designated object.
We use the Intersection Over Union (IoU) for that objective.
Several configuration are tested in an ablation study manner to decide the type and number prompts to optimize the performance of the SAM2 detections. 
We firstly experiment with the results of SAM2 without any prompts. We secondly introduce the prompts proposed by DINO networks: the center of the attention map and the different prompt values presented in Fig.~\ref{framework}

\begin{figure}[h]
        \centering
        \includegraphics[scale = 0.4]{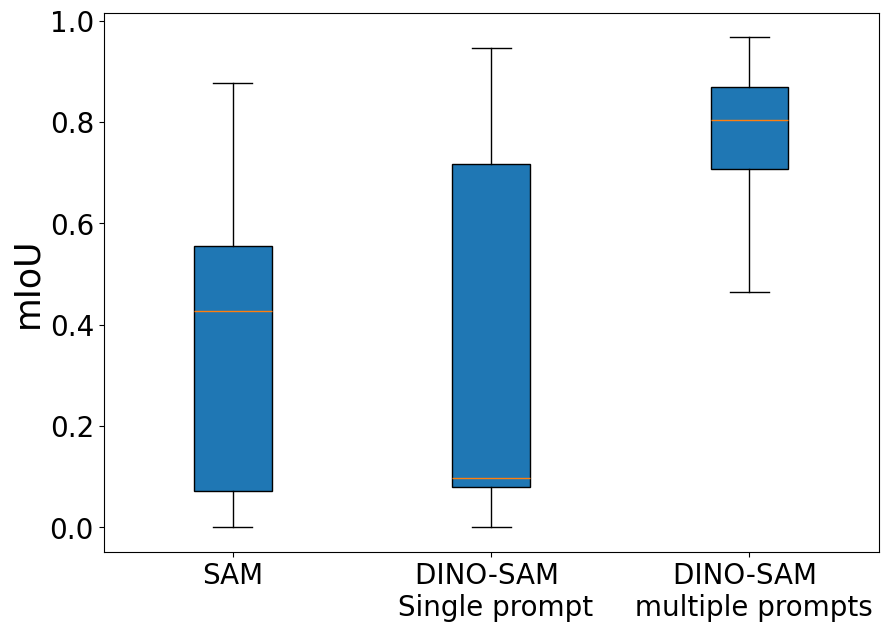}
        \caption{The value of mIoU of DINO-SAM for diffrent configurations and with different heuristics.}
        \label{fig:IoUSam}
    \end{figure}

The performance of solely using SAM is first evaluated, however, the model in this case failed to achieve reliable values of mIoU with only 35.37\%, with an upper quartile of 59\%.
Introducing DINO predicition with a single prompt ameliorated significantly the upper quartile without a significant impact on the value of mIoU that is 34.82\%.
On the other hand, after applying some heuristics such as clustering the thresholded mean attention map into different clusters, applying DBSCAN to propose a best fit bounding box and filtering the propmt points increased significantly the values of mIoU. After applying these heuristics, we obtained an mIoU of 75.26\%, with an upper quartile of 85\%. Fig.~\ref{fig:IoUSam} depicts experimenting with different heuristics and different configurations to retreive the mask of the object.

mIoU measures the similarity between the predicted segmentation mask and the ground truth mask by calculating the ratio of their intersection to their union.

The student model represented by the VAE architecture is lastly trained on a 32GB Tesla V100s GPU.
We train the model for 100 epochs before evaluating its performance with respect to manually labeled data.

\noindent\textbf{\textit{Validation metrics.}} Our validation metric consist of the euclidean distance between the ground truth labels and the centers of predicted heatmap of the student network. 
We denote the average value of  euclidean distance over the dataset, and for all the heatmaps as \textbf{\textit{$mED$}}. And the average euclidean distance for each heatmap as \textbf{\textit{$mED_k$}}. Ideally, this distance should not surpass 15 pixels.

\textbf{\textit{Baselines.}} We compare our method with the most relevant works found in the state of the art for similar type of problems such as using active contours~\cite{c201}, Kernelized correlation filters (KCF)~\cite{c205}, fully connected Convolutional Neural Networks~\cite{c203}, and also without the use of foundation models ~\cite{last}. 
Table ~\ref{tab:validation_baseline} shows the results of experiments and comparison with different baselines.

\begin{table}[h]
\centering
\caption{Mean Euclidean distance using different baselines.}
\label{tab:validation_baseline}
\scalebox{0.85}{\begin{tabular}{ccccccc}
                                                                           & \textit{\textbf{mED1}} & \textit{\textbf{mED2}} & \textit{\textbf{mED3}} & \textit{\textbf{mED4}} & \textit{\textbf{mED}} & \textbf{\begin{tabular}[c]{@{}c@{}}Standard\\ Deviation\end{tabular}} \\ \hline
\textbf{Active  Contour}                                                   & 141.70                 & 63.84                  & 42.74                  & 94.68                  & 85.74                 & 23.67                                                                 \\
\textbf{KCF}                                                               & 37.20                  & 83.51                  & 53.05                  & 38.5                   & 53.08                 & 30.62                                                                 \\
\textbf{SIFT}                                                              & 32.05                  & 26.87                  & 35.32                  & 36.97                  & 32.8                  & 3.85                                                                  \\
\textbf{\begin{tabular}[c]{@{}c@{}}CPD + Active\\ \\ Contour\end{tabular}} & 30.59                  & 30.74                  & 24.36                  & 32.73                  & 29.61                 & 9.84                                                                  \\
\textbf{CPD}                                                               & 33.77                  & 23.30                  & 24.34                  & 24.96                  & 26.59                 & 8.62                                                                  \\
\textbf{CNN}                                                               & 26.29                  & 16.42                  & 12.75                  & 9.59                   & 18.92                 & 12.99                                                                 \\
\textbf{VAE}                                                               & 9.55                   & 15.62                  & 13.07                  & 19.96                  & 14.55                 & 3.4                                                                   \\
\textbf{Ours}                                                               & 6.01                   & 15.35                  & 12.89                  & 19.34                  & 13.4                  & 3.1                                                                  
\end{tabular}
}\end{table}

\section{Discussion}

\noindent\textbf{\textit{Qualitative evaluation.}} As opposed to baseline methods, the student-teacher framework of our approach provided stable estimation of the shape of the deformable object without the need of any prior labeling. 
However, the automatically generated masks by the DINO-SAM2 framework were in some cases subject to ambiguity coming from erroneous prompt points, resulting low values of IoU.
This ambiguity is due to the presence of impurities and intrusions in some cases, such as the presence of greases on the robot trajectory.

\noindent\textbf{\textit{Quantitative evaluation}} In addition to qualitative evaluation, the calculated value of the Euclidean distance between the predicted points and the ground truth is relatively small and acceptable for such an application. 
However, this error is the result of two main factors:  human factor and the lack of visual features in some images. 
We recall that validation images were labeled manually with the possible centers of the heatmap,  which makes these annotations biased and prone to error. Since defining accurate labels for these tasks is not possible. To resolve this kind of uncertainty, we recommend adopting the automatically generated GT labels as ground truth for evaluation and not the manually defined GT.
As a proposed solution to this, we propose the use of sequential generative models students, such as Long Shot Term Memory Variational Autoencodre (LSTM-VAE)~\cite{lstmvae} capable of preserving temporal information about the process.

\noindent On the other hand, the lack visual features in some frames, due to several factors; such as noise and process instability have also resulted low fidelity predicitions of the heatmap.  

\section{Conclusion}
A clear representation of the shape of the manipulated object is a required step for robotic manipulation task.
In the case of deformable objects, this step is dependent on the shape of the deformations of the object. 
In this paper we address the case of highly-deformable fluid and viscoelastic objects.
We propose a teacher-student framework that does not require any prior labeling to represent them.
Our method is independent from the complex physical modeling requirements. 
During inference time, we use only the student of our framework. 
The student model, a generative model in our case, generates a heatmap that estimates the current state of the object.
This heatmap represents the locations at which the most important deformation occurs, according to the contour of the object.
The student network distils the knowledge of a teacher that is composed of two foundation models and the heatmap proposal algorithm~\ref{alg:normal_lines}.
The two foundation models employed in the case of this paper are DINO and SAM2. We use DINO as a prompts generator for the SAM2 model. 

We demonstrate first the capabilities of foundation models to retrieve pixel-level information about the deformable objects.
Secondly, we show that the generative student network is capable of learning the shape of the fluid deformable object by absorbing the knowledge provided by the foundation models successfully.

Further research directions include ameliorating the performance of the teacher by including the human in the loop through techniques such as active learning on one hand. On the other hand, integrating the retrieved shape of the weld pool in a control tasks such as visual servoing is also a perspective to this work.

\addtolength{\textheight}{-0cm}   %

\end{document}